\documentclass[10pt,twocolumn,letterpaper]{article}

\usepackage{cvpr}              

\definecolor{cvprblue}{rgb}{0.21,0.49,0.74}
\usepackage[pagebackref,breaklinks,colorlinks,allcolors=cvprblue]{hyperref}
\usepackage{pifont}
\usepackage{multirow}

\def\paperID{6995} 
\def\confName{CVPR}
\def\confYear{2025}

\title{CheckManual: A New Challenge and Benchmark for \\ Manual-based Appliance Manipulation}

\author{
Yuxing Long\textsuperscript{1,2},
Jiyao Zhang\textsuperscript{1,2},
Mingjie Pan\textsuperscript{1,2},
Tianshu Wu\textsuperscript{1},
Taewhan Kim\textsuperscript{1,2} and
Hao Dong\textsuperscript{1,2\dag}\\
\textsuperscript{\rm1}CFCS, School of Computer Science, Peking University \;
\textsuperscript{\rm2}PKU-Agibot Lab
}

\begin{document}
\maketitle
\begin{abstract}
Correct use of electrical appliances has significantly improved human life quality. Unlike simple tools that can be manipulated with common sense, different parts of electrical appliances have specific functions defined by manufacturers. If we want the robot to heat bread by microwave, we should enable them to review the microwave’s manual first. From the manual, it can learn about component functions, interaction methods, and representative task steps about appliances. However, previous manual-related works remain limited to question-answering tasks while existing manipulation researchers ignore the manual's important role and fail to comprehend multi-page manuals. In this paper, we propose the first manual-based appliance manipulation benchmark CheckManual. Specifically, we design a large model-assisted human-revised data generation pipeline to create manuals based on CAD appliance models. With these manuals, we establish novel manual-based manipulation challenges, metrics, and simulator environments for model performance evaluation. Furthermore, we propose the first manual-based manipulation planning model ManualPlan to set up a group of baselines for the CheckManual benchmark. Our project page is available at \href{https://sites.google.com/view/checkmanual}{https://sites.google.com/view/checkmanual}.

\end{abstract}  
\vspace{-6pt}
\section{Introduction}
\label{sec:intro}
Since the twentieth century, the emergence of appliances has significantly changed human society. In daily life, we extensively use various electrical appliances to assist our living and working. In recent years, robots have demonstrated enormous potential and value in areas like component assembly~\cite{luo2024hilserl}, goods sorting~\cite{fang2023anygrasp}, and indoor navigation~\cite{long2024instructnav}. Imagine if we could instruct robots to operate appliances to serve us, such as baking cakes with ovens, washing clothes with washing machines, and making juice with blenders, we would be further liberated from tedious household chores and have more free time.

\begin{figure}[htp]
    \centering
    \includegraphics[width=\linewidth]{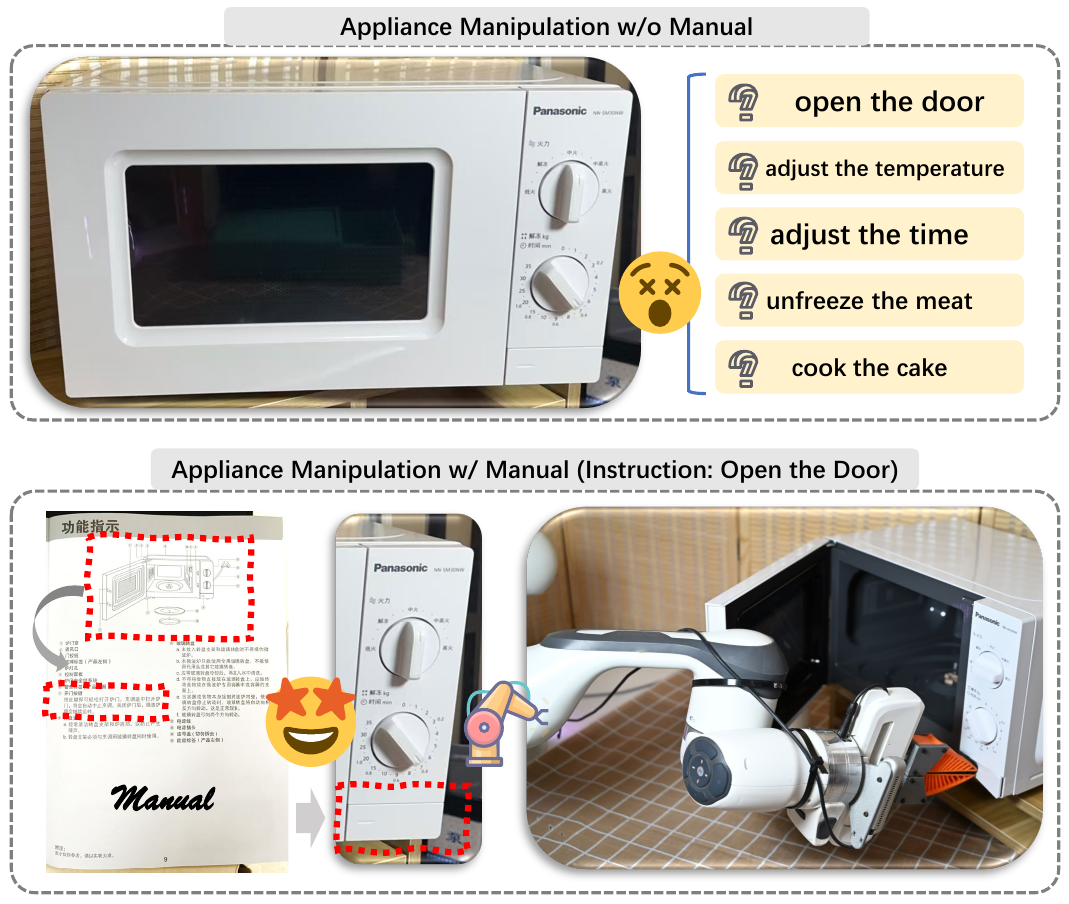}
    \vspace{-15pt}
        \caption{\textbf{Comparison between appliance manipulation with and without manual.} If the manual is absent, the concrete functions of the appliance part are unknown. As shown in the top image, the robot feels confused about tasks such as opening the microwave door. To correctly use the appliance, the robot must comprehend the manual content and follow its guidance to work.}
    \label{fig:Teasor}
    \vspace{-19pt}
\end{figure}

However, appliances are different from simple tools like hammers or teapots that can be correctly used based on common sense. One appliance may have multiple movable components and each component has specific functions even if they are the same type. Taking a microwave as an example, one button may control the temperature while the other button may set the time. Such function information is difficult to infer according to the appliance's visual appearance or human common sense. This time, we need to check the user manuals to learn about component functions and supporting tasks. Existing manipulation models like~\cite{brohan2023rt1}\cite{brohan2023rt2}\cite{kim24openvla}\cite{huang2023voxposer}\cite{huang2024rekep} already can adopt simple tools to do easy tasks but cannot read multi-page manuals not to say using the appliances to complete long-horizon complex tasks. Besides, previous manual-related research and academic datasets~\cite{nandy-s10}\cite{zhang2023mpmqa} are all collected by the NLP researchers\footnote{Note that the copyright of real manual belongs to the company. Using content from the manual for academic research requires authorization.}, which only focuses on page retrieval and question-answering tasks rather than manipulation. These academic datasets manual datasets also do not provide appliance CAD models and cannot be used for quantitative manipulation evaluation in the simulator. Therefore, there are no manual-based manipulation evaluation datasets and benchmarks, nor manipulation models capable of following manuals to use appliances at present.

To address this problem, we propose the first manual-based manipulation benchmark CheckManual. In the benchmark, we collect new evaluation data, design novel manipulation challenges, and propose new baseline models. Specifically, we first design an LLM-assisted human-revised manual creation pipeline based on appliance CAD models. For every CAD model, this pipeline can follow our surveyed real manual analysis to generate diverse appliance manuals with different part functions, manipulation demonstrations, and page layouts. Then, we set up three manual-based appliance manipulation challenge tracks including manipulation planning, CAD-assisted manipulation, and CAD-free manipulation. They progressively cover the key steps in manual-based appliance manipulation, comprehensively evaluating the model’s ability to operate appliances with manuals. Furthermore, we propose a manual-based manipulation planning method ManualPlan, and create a set of CAD-assisted primitive actions for appliance manipulation. The ManualPlan aligns manuals with appliances and plans detailed manipulation steps according to manual content, which can control primitive actions or open-vocabulary manipulation models to physically interact with appliances. These also act as the baseline methods for our CheckManual benchmark.

In this paper, our main contributions are:
\begin{itemize}
    \item[$\bullet$] We collect the first CAD model-aligned appliance manual dataset CheckManual through our devised large model-assisted pipeline. Every data collection stage undergoes careful human verification to guarantee correctness. 
    \item[$\bullet$] We design novel challenges for manual-based appliance manipulation considering different difficulty levels and conditions. Furthermore, we establish the evaluation environment in the simulator for the benchmark.
    \item[$\bullet$] We propose the first manual-based manipulation planning model ManualPlan to set up a group of baselines for three-level challenge tracks in CheckManual.
\end{itemize}

\vspace{-5pt}
\section{Related Work}
\subsection{Manipulation Large Models}
Large models pre-trained on Internet-scale text and image data emerge powerful capabilities for task planning and common sense reasoning. Therefore, leveraging large models has become a hotspot in embodied areas such as autonomous driving~\cite{yang2023lidar}, navigation~\cite{long2024discuss}~\cite{cai2024bridging} and manipulation. According to training and inference forms, existing manipulation methods can be divided into two main categories. The first category of work focuses on fine-tuning large models with manipulation trajectories collected from the simulator environment and real-world. The representative models are Google RT-series from RT-1~\cite{brohan2023rt1}, RT-2~\cite{brohan2023rt2}, and RT-Trajectory~\cite{gu2023rttrajectory} to STAR-RT~\cite{leal2024sarart}. These models take vision observation, robot states, and task instruction as input, and directly predict the low-level actions by multimodal large models. Similar open-sourced work includes ManipLLM~\cite{manipllm}, RoboFlamingo~\cite{li2023vision} and OpenVLA~\cite{kim24openvla}. The other category of work makes efforts to effectively integrate large models' zero-shot inferences into the manipulation planning and decision process. SayCan~\cite{ahn2022can} first utilizes a large language model to score a group of kills to make progress toward the high-level task. Voxposer~\cite{huang2023voxposer} drives a multimodal large model to create 3D value maps to guide the robot toward the target object part. CoPa~\cite{huang2024copa}, ReKep~\cite{huang2024rekep}, and OmniManip~\cite{omnimanip} leverage multimodal large models to generate spatial-temporal constraints for complex tasks like ``\emph{insert the flower into vase}'' or ``\emph{pour tea into water}''. However, all these large model-based methods can only make plans based on their internal common sense. Considering heating the bread with the microwave, they even struggle to figure out the temperature knob and timer knob, not to say complete the whole task.

\begin{figure*}[htp]
	\centering
	\includegraphics[width=\linewidth]{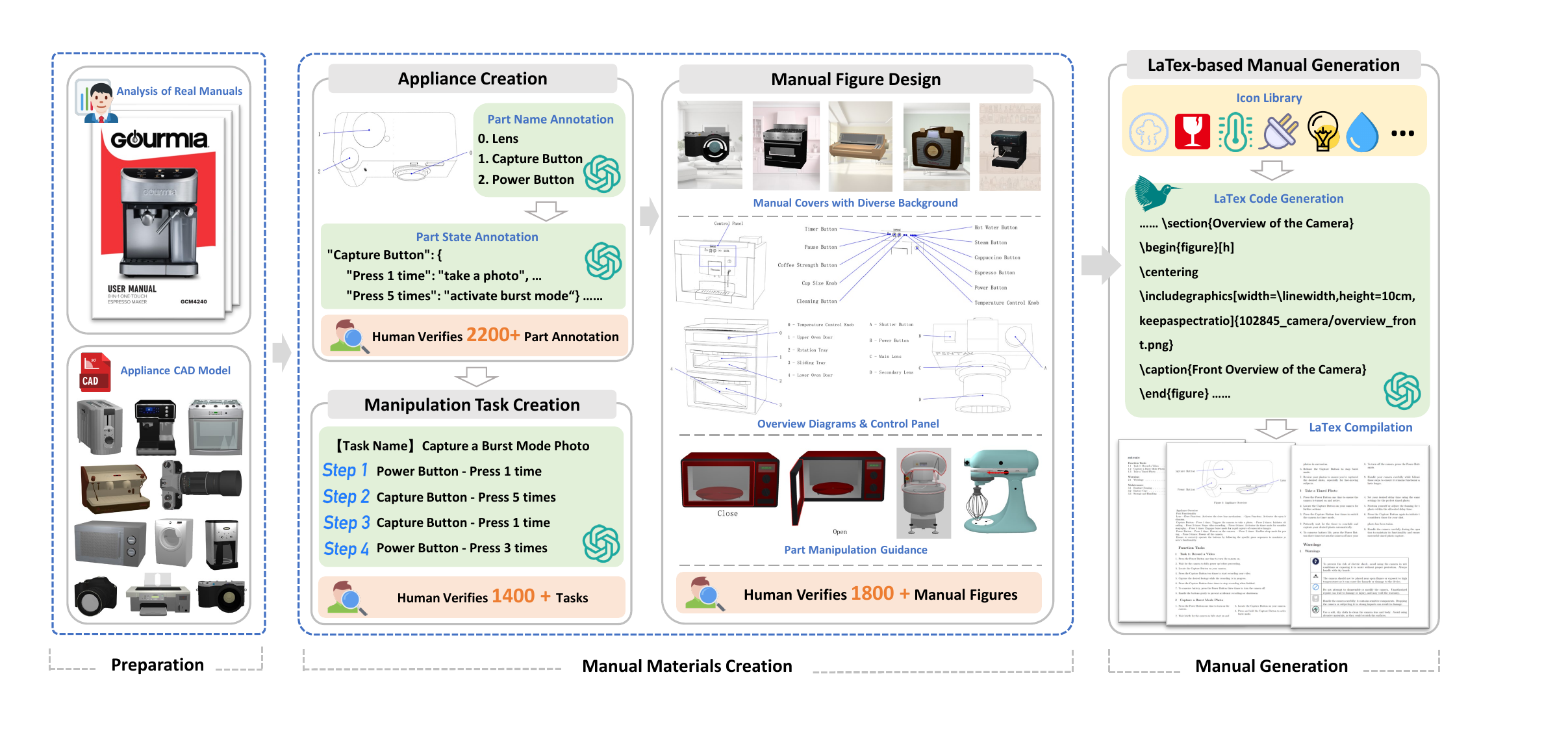}
	\vspace{-15pt}
	\caption{\textbf{Generation workflow of CheckManual dataset.} In the leftmost part (Section~\ref{sec:preparation}), we analyze real manuals to learn about their formats and collect different categories of appliance CAD models. The middle part (Section~\ref{sec:appliance_creation} - \ref{sec:manual_figure}) demonstrates the creation of manual materials, including appliance creation, task generation, and figure design. Human verifies every step to guarantee correctness. Based on this information, the rightmost part (Section~\ref{sec:manual_gen}) generates appliance manuals with diverse formats through the \LaTeX{} codes. }
	\label{fig:CheckManual}
	\vspace{-10pt}
\end{figure*}

\subsection{Appliance Manual-based Tasks}
NLP researchers have recognized the manual's importance in addressing user questions. TechQA~\cite{nandy-techqa} collects 1400 user questions from online forums and annotates the corresponding answers from IBM technical documents to set up \emph{Document-based Question Answering} task. S10 QA~\cite{nandy-s10} manually creates 904 QA pairs based on the manual of the Samsung S10 phone. Nandy et al.~\cite{nandy-s10} also collect 950 QA pairs based on the Samsung Smart TV manual to generate Smart TV/Remote QA dataset. These two datasets are utilized to establish a \emph{Manual-based Question Answering} benchmark. The above datasets are all pure text corpus. PM209~\cite{zhang2023mpmqa} collects 209 product manuals from 27 well-known consumer electronic brands to set up \emph{Manual Page Retrieval} task and \emph{Manual-based Multimodal Question Answering} task. Although these datasets and tasks are related to the appliance manual, they are all designed for NLP evaluation and lack corresponding appliance CAD models. Therefore, manual-based manipulation benchmarks cannot be directly established on these existing related researches. We create the first manual \& CAD model-aligned dataset and design novel benchmarks and baseline models for manual-based appliance manipulation.

\section{Manual-based Manipulation Benchmark CheckManual}
In this part, we will introduce the creation process of our CheckManual dataset in Section~\ref{sec:manual_creation}, analyze the numerical and visual statistics about CheckManual in Section~\ref{sec:statistic}, and define the evaluation for robot manipulation based on these manuals in Section~\ref{sec:evaluation}.

\subsection{Appliance Manual Creation}
\label{sec:manual_creation}
To create an open-source available manual dataset, We design a large model-driven human-verified manual generation workflow based on appliance CAD models. Figure~\ref{fig:CheckManual} demonstrates the generation workflow of the CheckManual dataset, which will be detailed in the following.

\subsubsection{Preparation}
\label{sec:preparation}
Before formally starting manual creation, we analyze a stack of real appliance manuals to learn about real manual formats and collect available articulated appliance CAD models for manual creation.

\noindent\textbf{Analysis of Real Appliance Manuals}
 We collect 110 appliance manual PDFs from the Internet, which cover 11 categories of our dataset involved appliances. These manuals come from different countries and regions, including China, the European Union, Japan, and India. Then we manually analyze each manual from \emph{Functional Part Annotation Type}, \emph{Guidance of Functional Part Operation}, and \emph{Task Expression Format} these three aspects. We find that the functional parts are usually explained in the point-line annotation formats. These lines connect the points on appliance parts and the function names or IDs of parts as shown in the middle of Figure~\ref{fig:CheckManual}.~\footnote{We use examples from our ChcekManual to avoid any infringement} The guidance of functional part operation is commonly represented in formats: \ding{182} Pure text description like ``\emph{Rotate temperature knob 90° clockwise}''. \ding{183} Close-up figures of the manipulated part, such as the close and open states of the red microwave door in Figure~\ref{fig:CheckManual}. \ding{184} Part motion decomposition diagram that shows the dynamic change of part during its motion process. Figure~\ref{fig:CheckManual} shows one example of opening the white mixer lid. \ding{185} Motion trajectory of the contact point on the part, like manipulating Figure~\ref{fig:CheckManual}'s blue mixer slider from right to left. Besides, the manipulation tasks in the real manuals are often depicted in the text, bullet lists, ordered lists, and tables with one-column or multi-columns.

\noindent\textbf{Appliance CAD Models}
Our appliance CAD models come from the PartNet-Mobility~\cite{Mo_2019_CVPR} Dataset, which contains common categories of household appliances CAD models, including microwaves, dishwashers, and refrigerators. Appliance models in this dataset are annotated with manipulable parts, such as buttons, knobs, levers, sliders, and so on. In total, we collect 11 categories, 182 CAD models, which cover widely-used household appliances in our daily lives.

\begin{figure*}[htp]
	\centering
	\includegraphics[width=\linewidth]{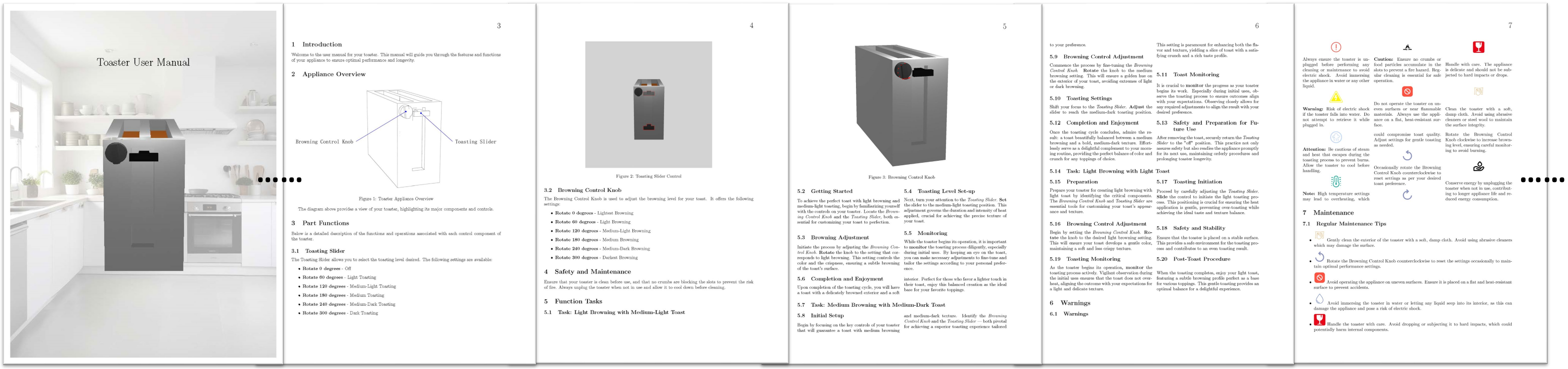}
	\vspace{-0.3cm}
	\caption{\textbf{One example of appliance manual from our CheckManual dataset.} Please zoom in for details.}
	\label{fig:ManualCase}
	\vspace{-0.3cm}
\end{figure*}

\subsubsection{Appliance Creation}
\label{sec:appliance_creation}
With the appliance CAD models, we can create concrete appliances by annotating each movable part with a specific function name (\emph{e.g., temperature knob}) and corresponding function states (\emph{e.g., clockwise 60° -- 120℃ and clockwise 120° -- 160℃}). This process is detailed in the following.

\noindent\textbf{Part Function Name Annotation}
Considering part function names are closely related to the appliance layout and parts' visual features, we drive the annotation process by the multimodal large language model (MLLM). Before annotation, manipulable part IDs and types are automatically labeled on the appliance overview figures through point-line annotation. Taking these figures as visual prompts, we instruct the MLLM to design the function name of each part. The multimodal large model should consider the appliance type and carefully analyze part visual information to create reasonable annotations. Re-generation mechanism would be triggered when the number of annotated parts is unequal to the total number of manipulable parts. By resolving the large model's prediction with regular expressions, we can obtain a dictionary recording the one-to-one relationships between part ID and corresponding function name. One such dictionary can be used to generate one appliance. We can create multiple appliances from one CAD model by repeating the above annotation process.

\begin{figure*}[htp]
	\centering
	\includegraphics[width=\linewidth]{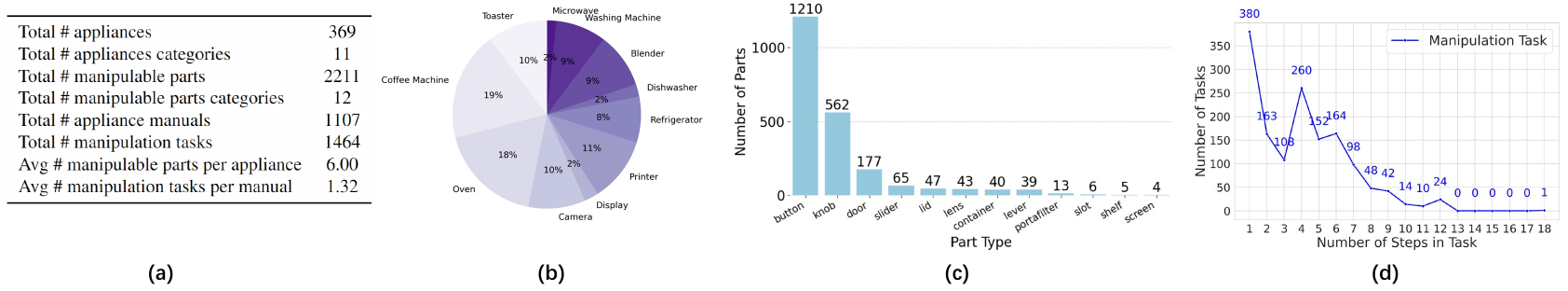}
	\vspace{-0.5cm}
	\caption{\textbf{CheckManual dataset statistics.} The above tables and figures from left to right are (a) Numerical statistics. (b) Proportion of different appliances. (c) Distribution of manipulable components on the appliances. (d) Distribution of manipulation task step lengths.}
	\label{fig:statistic}
	\vspace{-0.3cm}
\end{figure*}
\noindent\textbf{Part Function State Annotation}
To annotate part function states, we first sample available states for each manipulable part according to their types. For buttons, we randomly sample available push times in a pre-defined range. For knobs and levers, we randomly sample several rotation degrees at intervals of 60°. Other manipulable parts like doors, lids, and containers, are assigned with ``open'' and ``close'' states. With part function names and available states, we instruct the large language model (LLM) to design concrete functions for each state. This way, each manipulable part on the appliance will be annotated with multiple function states.

\noindent\textbf{Human Verification about Part Functions}
Although large models are powerful annotation tools, they still make intolerable mistakes and thus require human verification. In the annotation about function names, MLLM sometimes fails to identify tiny visual elements on the CAD models like already-drawn button logos or scales, which may lead to mismatching between annotated function names and part visual appearance. In the annotation about function states, LLM may create an anti-commonsense design such as ``500℃ for oven temperature knob''. Therefore, we check large model-annotated function names and states through our designed human verification website (See Appendix) to revise incorrect cases. In this stage, we manually verify more than 2200 part function annotations.

\subsubsection{Manipulation Task Creation}
\label{sec:task_creation}
The manual often includes several representative tasks to introduce how to use the appliance. These tasks are also the core of our manual-based manipulation evaluation. We can create appliance-executable tasks based on their part function names and states.

\noindent\textbf{Task Proposal}
With function names and states of movable parts on the appliances, we instruct the LLM to propose short-horizon and long-horizon manipulation tasks for each appliance. Every task step should specify the part function name and the manipulated state in a resolvable format like ``\emph{1. Front Door, Open}'', ``\emph{2. Power Button, Push 1 time}'' ... To improve task quality and reduce repetition, the LLM only needs to design one task in one inference based on already created tasks.

\noindent\textbf{Human Verification about Manipulation Tasks}
In the task creation, LLM may also make anti-commonsense mistakes. To guarantee absolute correctness for evaluation, after the LLM completes task generation, we manually check every task step to adjust unreasonable step orders and revise incorrect steps. If the task is completely unreasonable, we will manually create a new one from scratch. Totally, we checked over 1400 manipulation tasks created by LLM.

\subsubsection{Manual Figure Design}
\label{sec:manual_figure}
According to our analysis of real manuals, cover, control panels, and part manipulation guidance are the main figures in the manuals. Therefore, we devise methods to create these figures.

\noindent\textbf{Manual Cover}
The manual cover usually displays appliance appearance and typical usage scenes. we predefined a set of candidate camera poses to capture the appliance from different views. The views with the largest total areas of manipulable parts are selected to render RGB appliance overview figures. To synthesize custom usage scenes, we further add the indoor background to the appliance overview through text-to-image generation model SDXL-Lightning~\cite{lin2024sdxllightning}. The background room type is chosen based on the appliance category. For example, the oven is usually put into the kitchen while the washing machine is often arranged in the laundry room. 

\noindent\textbf{Overview Diagram \& Control Panel}
The overview diagram and control panel demonstrate the function name of every movable part on the appliance. Before annotation, we first apply the DBSCAN~\cite{ester1996density} algorithm to try clustering buttons and knobs based on their geometric centers. If clustered buttons and knobs are found, the areas with them are selected as control panels and cropped from the appliance overview figures as close-ups of control panels. Then, we adopt point-line annotation methods (Figure~\ref{fig:CheckManual}) from the real manuals to draw overview diagram and control panel based on part function names created in the previous stage. Furthermore, considering real manuals often deliver overview diagrams and control panels in line drawings, we create sketch-style figures by conducting edge detection on the RGB normal maps. This way, we can generate these two kinds of figures both in RGB style and sketch style.

\noindent\textbf{Part Manipulation Guidance}
This kind of figure usually introduces the physical interaction approach for appliance manipulable parts. To generate such figures, we first render manipulable parts according to their types and visible surface areas. For buttons, knobs, levers, and drawers, we render them with the camera axis parallel to their joint axes. Other manipulable parts, including doors, sliders, lids, and containers, are rendered with the camera axis perpendicular to their joint axes while ensuring maximum visible areas. Then, we draw manipulation guidance on these figures to display how to manipulate them. Following common representation in the real manuals from Section~\ref{sec:preparation}, we adopted these four different strategies to draw manipulation guidance as shown in Figure~\ref{fig:CheckManual}. To realize this, we manually annotate the 3D position of one contact point for every manipulable part on all appliance CAD models and calculate its motion trajectory according to the part-linked joint axis.

\noindent\textbf{Human Verification about Manual Figures}
When collecting appliance overviews based on the largest visible part areas, the view may not be suitable to observe every part. Besides, sketch-style figures may contain missing lines or abnormal shadows. Therefore, we conduct human verification on more than 1800 figures to trigger regeneration when these errors occur.

\begin{table}[h]
	\centering
    \caption{\textbf{Comparison of different manual-based manipulation challenge tracks in our CheckManual benchmark.}}
	\scalebox{0.46}{
		\begin{tabular}{c|c|c|c|c|c}
			\toprule
                \multicolumn{4}{c|}{\textsc{\textbf{Available Input}}} & \multicolumn{2}{c}{\textsc{\textbf{Prediction Target}}} \\
                \midrule
			\textbf{Task Instruction} & \textbf{RGB-D} & \textbf{Manual PDF} & \textbf{Appliance CAD Model} & \textbf{High-level Planning} & \textbf{Low-level Manipulation} \\
                \midrule
                \multicolumn{6}{c}{\textbf{Track 1: Manual-CAD-Appliance Aligned Planing}} \\ 
			\midrule
		      \checkmark & \checkmark & \checkmark & \checkmark &\checkmark  &  \\
                \midrule
                \multicolumn{6}{c}{\textbf{Track 2: Manual \& CAD based Manipulation}} \\ 
			\midrule
			\checkmark & \checkmark & \checkmark & \checkmark &  & \checkmark \\
                \midrule
                \multicolumn{6}{c}{\textbf{Track 2: Manual \& CAD based Manipulation}} \\ 
			\midrule
			  \checkmark & \checkmark & \checkmark &  &  & \checkmark \\
			\bottomrule
		\end{tabular}
	}
    \vspace{-0.3cm}
	\label{tab:tracks}
    \vspace{-0.4cm}
\end{table}

\subsubsection{\LaTeX{}-based Manual Generation}
\label{sec:manual_gen}
As we know, \LaTeX{} is a document programming language that offers both the advantages of standardization and flexibility. We can not only keep the same document style from page to page but also easily insert different types of figures and tables into the document by \LaTeX{} coding. Recently, large language models emerged with powerful capabilities for code generation. Therefore, we decided to leverage LLM to generate the \LaTeX{} codes for the manual and then compile them into a PDF format appliance manual as shown in the rightmost part of Figure~\ref{fig:CheckManual}. 

Before starting this process, we pre-defined a set of widely used \LaTeX{} packages (\emph{e.g., multirow, xcolor, caption}) and prepared a collection of commonly used icons (\emph{e.g., fragility reminder, temperature warning, environment protection}). To increase diversity, we randomly sample different formats of covers, diagrams, and guidance demonstrations from the Section \ref{sec:manual_figure} as figure resources and randomly select a text style from real manuals (\emph{e.g., text, bullet list, table}) for task description. In the generation, we synthesize the prompt with textual part function annotation, manipulation tasks, and above image paths to require the large model to continue one section of \LaTeX{} codes based on the previous. To prevent uncompilable codes, we design a section-granularity compilation mechanism. The newly generated section of \LaTeX{} codes is appended to the end of the previous correct code and compiled together. Regeneration will be triggered when errors occur in the code compilation. Figure~\ref{fig:ManualCase} displays one manual from our CheckManual dataset.

\subsection{Dataset Statistic Analysis}
\label{sec:statistic}
We totally created 1107 manuals for 369 appliances, which cover 11 common appliance categories in daily life, including washing machines, coffee makers, dishwashers, microwaves, and so on. These appliances contain 12 types, with 2211 different manipulable parts. On average, there are 6.00 manipulable parts on each appliance. In terms of quantity and diversity, parts contained in our dataset can support most appliance operations. Besides, our dataset includes 1464 manipulation tasks in total. Every task is accompanied by step-by-step manipulation parts and target states. These tasks include both single-step and long-horizon tasks with a maximum of 18 steps. The table and charts in Figure~\ref{fig:statistic} display comprehensive statistics about our CheckManual dataset.


\subsection{CheckManual Challenge Tracks}
\label{sec:evaluation}
With the collected manual data, we design challenge tracks for manual-based manipulation. The basic setting is given a task instruction $T$ and a corresponding appliance PDF manual $M$, the robot should follow the manual to complete the tabletop appliance manipulation task with RGB-D observation $I_c$ and $I_d$. Based on this basic setting, we design the challenge tracks as shown in Table~\ref{tab:tracks} and the following.

\begin{figure*}[htp]
	\centering
	\includegraphics[width=\linewidth]{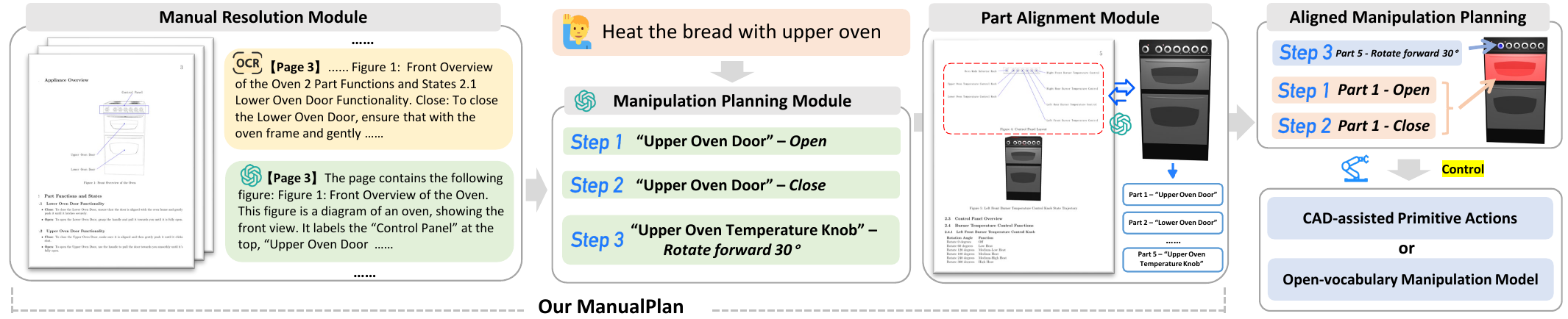}
	\vspace{-0.3cm}
	\caption{\textbf{The framework of our ManualPlan model.} It is composed of Manual Resolution, Manipulation Planing and Part Alignment modules. The ManualPlan can make high level planning to control the CAD-assisted primitive actions or open-vocabulary manipulation large model (\emph{e.g.}, VoxPoser~\cite{huang2023voxposer}) to use the appliance, which serves as the baseline models for CheckManual benchmark.}
	\label{fig:ManualPlan}
	\vspace{-0.3cm}
\end{figure*}

\noindent\textbf{Track 1: CAD-Appliance Aligned Planing}
We first consider an ideal setting where the appliance CAD models are available and CAD-Appliance is aligned. The ”CAD-Appliance
Aligned” means part area and ID are accessible on the camera observation. In this case, the evaluation focuses on testing the robot's capability of analyzing appliance manuals and making manipulation plans based on the given task instruction and manual content. The evaluation is composed of two aspects. Firstly the robot should align all function names in the manuals with appliance parts. Secondly, the task planning should include correct manipulation target part and primitive actions in every task step. This challenge track is free from manipulation execution.

\noindent\textbf{Track 2: Manual \& CAD based Manipulation}
In this task, the appliance CAD models are available but the mapping relationships among parts on manual, CAD model, and observation are not known. This setting is reasonable because it is easy for manufacturers to provide plain CAD models for their appliance products. For evaluation, the robot should follow the given task instruction to execute low-level manipulation with a two-finger gripper based on manual content and CAD model. To complete this challenge, the robot needs to first estimate the 6-DoF pose of observed objects before using the CAD model. After that, the robot should also align parts on the manual, CAD model, and current observation. After completing the part alignment, the robot can obtain part-linked joint position and rotation axis from the CAD model, which will help the robot to make manipulation planning and execution.

\noindent\textbf{Track 3: Pure Manual-based Manipulation}
We further consider the setting that aligns with real-world applications where only appliance manuals are available to the robot. The challenges of this setting come from two aspects. On the high level, the robot should analyze the appliance manual to learn about the function information (\emph{i.e., name and possible states}) and manipulation methods of every movable part on the appliances. Then, the robot should plan the detailed manipulation steps for the given task based on the manual content and common sense. On the low level, the robot must execute the high-level plan step-by-step with its end-effector to complete the instruction-specified task. During the whole process, the robot can only rely on its RGB-D observation and manuals instead of any CAD-model assistance. For both Track 2 and Track 3, any procedural error during the operation will cause the entire task to fail. 

\noindent\textbf{Evaluation Metrics}
To evaluate model performances on the above challenge tracks, we define the task-level and step-level metrics. On the step level, we evaluate the alignment success rate and planning success rate for Track 1 and task execution success rate and average task completion rate for Track 2 and 3. The alignment success is recognized when all CAD model parts are linked to the correct function names mentioned in the manual. The task planning success is confirmed when part IDs and manipulation actions in every planned step are all correct. The task execution is regarded as a success when all task steps are completed in the correct order. The task completion rate is calculated by ``\emph{\# Success Steps / \# Task Steps}''. Note that only steps before the failure step can be counted as successful steps. We take the average values of the above metrics as final results.  On the step level, we check the state of the target part after each manipulation step to judge the success. Specifically, we define the whole movement limitation range of the target joint as $L$. For revolute parts like knobs, levers, and doors, the acceptable error range is defined as $L\;\pm\;30$°. For prismatic parts like the button, container, and slider, the successful manipulation is confirmed when the button moves $> 25\%\; L$ or ``\emph{$MD_{part}$ / $MD_{appliance}$}'' $> 50\%$ if target part and appliances moves in the same direction, where $MD$ represented motion distance.



\section{Manual-based Manipulation Baselines}
In this section, we design baseline models for the three tracks of the benchmark. For track 1, we propose the ManualPlan model, which focuses on manual-based task planning, as detailed in Section~\ref{sec:manualplan}. For tasks involving robotic execution, corresponding to track 2 and track 3, we extend the ManualPlan model by integrating it as a high-level planning module with existing manipulation models, as discussed in Section~\ref{sec:low-level_manipulation}.

\subsection{ManualPlan for High-Level Task Planning}
\label{sec:manualplan}
ManualPlan introduces a novel framework for high-level task planning of robotic manipulation with appliance manuals. It processes manual content, generates detailed plans, and aligns parts in the real appliance with those referenced in the manual.

\noindent\textbf{Manual Resolution Module.}
The appliance manual in our dataset is in PDF format, which is consistent with the real manuals. The PDF file cannot be directly processed by large language models. Therefore, we divide the manual PDF file into page-level images first. Then, we extract text content from these page images using the Paddle OCR model. We also utilize the multimodal large language model (MLLM) to detect visual elements like figures and tables on each page image. In this way, we can obtain the page-granularity text and vision information about the whole manual. 

\noindent\textbf{Manipulation Planning Module.}
With extracted manual information, we call LLM to plan a step-by-step appliance manipulation process for the given task. The large language model must specify the part function name and manipulated states in the dictionary format like ``\emph{\{'Brew Strength Lever': 'Rotate 60 degrees'\}}'' for every planning step. Only the manual-mentioned parts can be used in the plan. Considering just a part of the tasks are demonstrated in the manual, the LLM should make plans based on both common sense about appliance usage and the manual content. Any unresolvable prediction will trigger the regeneration request.

\noindent\textbf{Part Alignment Module.}
To achieve this goal, we incorporate a visual prompting mechanism known as Set-of-Mark (SoM)~\cite{yang2023setofmark} to align parts in the field of view and parts depicted in the manual. SoM leverages segmentation models to partition an image into distinct regions and assign a numeric marker to each, which can significantly improve the visual grounding capabilities of multimodal large language models. Here, we utilize the open-vocabulary object detection model Grounding-DINO ~\cite{liu2024groundingdino} to predict bounding boxes of movable parts and prompt Segment Anything~\cite{kirillov2023segany} with these bounding boxes to generate movable part masks. These masks are assigned different colors with transparency and overlayed on the observation image. The ID is marked at the geometric center of each part mask. Then, we provide the observation image and appliance diagrams from manual to MLLM requiring it to identify the mapping relationships from mask IDs to part function names. This way, we realize the part alignment between the observation and the manual. For the Manual \& CAD Model-based Manipulation task, although the CAD model is available, the part function names still need to be aligned with parts on the CAD model. Therefore, we use a similar SoM-based approach to achieve such alignment with MLLM. For the other two tasks, the CAD model is unavailable for the Pure Manual-based Manipulation task while part alignment is already achieved in the Manual-CAD-Appliance Aligned Manipulation task. \\

\subsection{ManualPlan-Driven Low-Level Manipulation}
\label{sec:low-level_manipulation}

\noindent When executing manipulation on the appliance, we employ ManualPlan as a high-level planner and design different manipulation policies for Track 2 (CAD Model Available) and Track 3 (CAD Model Unavailable).

\begin{table*}[tb]
    \begin{center}
    \small
    \caption{\textbf{Baseline model's performances on three challenge tracks of CheckManual benchmark.} The results of Track 1 are reported in the ``\emph{Part Alignment Success Rate / Task Planning Success Rate}'' format. For Track 2 and 3, their results are reported in ``\emph{Task Completion Rate / Task Success Rate}'' format. The ``\emph{w/o manual}'' means ablating appliance manuals input for the above methods.} 
    \scalebox{0.66}{
        \setlength{\tabcolsep}{1.0mm}{
        \begin{tabular}{c|c|c|c|c|c|c|c|c|c|c|c|c}
  \toprule
    \multirow{1}{*}{\centering \textbf{\textsc{Baseline Model}}}
    & \includegraphics[width=0.035\linewidth]{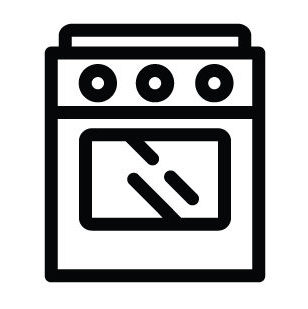}
    & \includegraphics[width=0.035\linewidth]{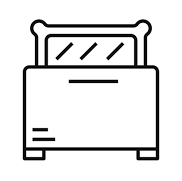}
    & \includegraphics[width=0.035\linewidth]{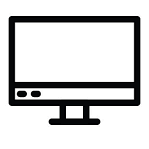}
    & \includegraphics[width=0.035\linewidth]{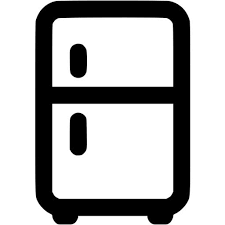}
    & \includegraphics[width=0.035\linewidth]{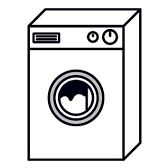}
    & \includegraphics[width=0.035\linewidth]{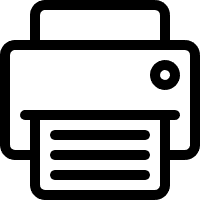}
    & \includegraphics[width=0.035\linewidth]{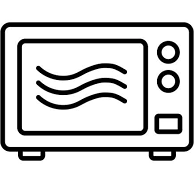}
    & \includegraphics[width=0.035\linewidth]{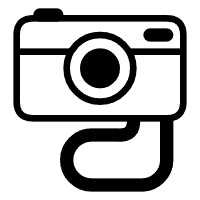}
    & \includegraphics[width=0.035\linewidth]{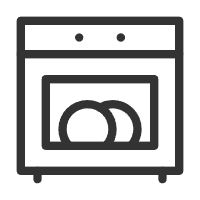}
    & \includegraphics[width=0.035\linewidth]{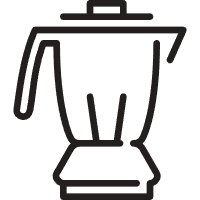}
    & \includegraphics[width=0.035\linewidth]{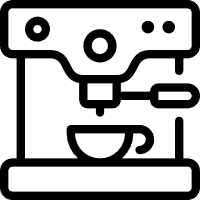}
    & \textbf{Total} \\ 
  \toprule
  \multicolumn{13}{c}{\textbf{Track 1: Manual-CAD-Appliance Aligned Planing}} \\ 
  \midrule
  \textbf{ManualPlan}  &22.96 / 22.22   &39.47 / 22.82    &18.52 / 14.29    &72.84 / 60.19  &4.17 / 3.91    &8.13 / 8.64    &94.44 / 68.00    &47.22 / 30.56    &41.67 / 40.62    &21.15 / 14.39   &15.46 / 9.96    &27.32 / 20.70            \\ 
  \quad w/o manual & -- / 4.21  & -- / 2.68   & -- / 20.00  & -- / 48.15  & -- / 0.78   & -- / 2.47   & -- / 36.00   & -- / 6.94   & -- / 15.62   & -- / 4.32   & -- / 2.85   & -- / 7.99 \\ 
  \midrule
  \multicolumn{13}{c}{\textbf{Track 2: Manual \& CAD based Manipulation}} \\ 
  \midrule
  \textbf{ManualPlan + Primitive Actions}  &2.39 / 1.15   &7.12 / 4.70  &13.63 / 11.43  &0.46 / 0.00   &0.78 / 0.78  &7.91 / 3.70  &10.00 / 4.00  &11.11 / 4.86  &3.12 / 0.00   &2.32 / 1.44  &2.56 / 1.42  &4.50 / 2.39   \\ 
  \quad w/o manual &1.05 / 0.38  &1.76 / 1.34  &21.39 / 14.29  &0.46 / 0.00   &0.78 / 0.78  &3.50 / 1.23  &2.00 / 0.00   &5.09 / 1.39  &1.04 / 0.00  &2.28 / 2.16    &1.69 / 1.42  &2.47 / 1.37  \\ 
  \midrule
  \multicolumn{13}{c}{\textbf{Track 3: Pure Manual-based Manipulation}} \\ 
  \midrule
  \textbf{ManualPlan + VoxPoser~\cite{huang2023voxposer}}  &6.11 / 1.92    &7.18 / 4.70  &7.10 / 5.71  &3.01 / 1.85   &3.12 / 3.12   &7.99 / 3.70    &0.00 / 0.00    &10.66 / 6.25    &6.75 / 3.12    &0.48 / 0.00  &2.96 / 2.14   &5.18 / 2.87        \\ 
  \quad w/o manual &0.99 / 0.38   &1.39 / 0.67    &0.57 / 0.00  &0.46 /0.00   &1.56 / 1.56   &1.54 / 0.62  &0.00 / 0.00  &2.50 / 0.69   & 0.00 / 0.00  &0.00 / 0.00  &0.98 / 0.71   &1.11 / 0.55             \\
  \bottomrule
        \end{tabular}}
    }
    \vspace{-0.3cm}
    \label{tab:result}    
    \vspace{-0.3cm}
    \end{center}
\end{table*}

\noindent\textbf{CAD-assisted Manipulation Policy.}
When the CAD model is available, we can estimate the 6-DoF pose of the observed appliance in the camera frame by the pose estimation models like~\cite{foundationposewen2024}\cite{zhang2024omni6dpose}\cite{zhang2023generative}. With such an estimated object pose, the observed appliance can be aligned with its CAD model and all joints' positions and axis directions on the observed appliance become available. Therefore, we design primitive actions for different appliance parts to synthesize CAD-assisted manipulation policy. For door and parts with slider (\emph{e.g., screen, slider, lid}), we provide the 3D bounding box of the target part to constrain the AnyGrasp~\cite{fang2023anygrasp} model to predict part-level grasp poses. Then, the gripper poses in the motion sequences can be calculated based on the information about the joint and manipulation method. For the button part, we take its surface geometrical center as the contact point and control the closed gripper to push the button from the normal direction of the contact point. For the knob part, we control the open gripper to get close to the knob from the normal direction of its surface geometrical center and close the gripper to grasp. After the gripper successfully grasps the knob, it will rotate planned degrees around the normal of its surface geometrical center. 

\noindent\textbf{CAD-free Manipulation Policy.}
We integrate a recently proposed open-vocabulary manipulation model VoxPoser~\cite{huang2023voxposer} into our ManualPlan to complete the CAD-free manipulation task. In this framework, ManualPlan and VoxPoser cooperate like the cerebrum and cerebellum in the human brain. ManualPlan reads manuals, makes high-level planning, and aligns observed components with manual-depicted parts. Then, it grounds the target component and demonstrates the corresponding manipulation method in language format, which is transferred to VoxPoser with camera RGB-D observation to execute low-level manipulation action on the appliance.
\section{Experiments}

\subsection{Implementation Details}
When creating the CheckManual dataset, we utilize GPT-4o~\cite{achiam2023gpt} as the large language model and multimodal large language model. Following previous work~\cite{vatmart}, we use the Franka Panda flying gripper in the simulation. The goundingDINO~\cite{liu2024groundingdino} model is utilized to ground the bounding boxes of appliance components. The predicted bounding boxes prompt SAM~\cite{kirillov2023segany} model to generate component masks. FoundationPose~\cite{foundationposewen2024} helps us estimate appliance 6-DoF pose in the camera observation. In the evaluation, manual comprehension and manipulation planning processes are assisted by GPT-4o~\cite{achiam2023gpt} model.

\subsection{Simulation Evaluation}
We evaluate our proposed ManualPlan model and ManualPlan-driven framework using the CheckManual benchmark, which employs SAPIEN~\cite{Xiang_2020_SAPIEN} as the simulation. This environment provides a realistic and flexible simulation for testing robotic manipulation tasks across a variety of appliances. The experimental results on different tasks and appliance categories are summarized in Table~\ref{tab:result}.
From the results of Track 1, we observe that the ManualPlan model is capable of generating a part of correct task plans based on multi-page manuals. However, even the most advanced Multi-Modal Large Language Models (MLLMs) still face significant challenges in aligning manual part annotations with the corresponding CAD models of appliances, which greatly limits the overall performance of ManualPlan in task planning.
For Track 2 and Track 3, the performance of the ManualPlan-driven framework highlights the difficulties posed by long-horizon, manual-based manipulation tasks. After analyzing failure cases, we identified that the accumulation of errors in part identification, task planning, and multi-step execution are the primary causes of failure. 
Additionally, we conducted an ablation study by removing the manual (\emph{i.e., w/o manual}) to investigate its impact on task performance. As shown in Table~\ref{tab:result}, the manual plays a crucial role in guiding the robot to accurately plan and execute manipulation tasks, underscoring its importance in completing appliance manipulation tasks.

\subsection{Real Robot Deployment}
To demonstrate the potential of our approach in real-world applications, we conducted real-world robot experiments following the task definition of Track 3 from the Check-Manual benchmark. The hardware setup involved a Franka Panda robot arm and a RealSense D415 RGB-D camera mounted on the end-effector. Considering error accumulation in the long-horizon execution, we follow~\cite{huang2024rekep} to cache correct manipulation actions in every step.

\section{Conclusion}
In this work, we introduce CheckManual, the first benchmark for manual-based appliance manipulation, and define new challenges, metrics, and simulator environments for model evaluation. Additionally, we present the first manual-based manipulation planning model ManualPlan to set up a group of baselines for the CheckManual benchmark.
\section*{Acknowledgment}
This work was supported by the National Youth Talent Support Program (8200800081) and National Natural Science Foundation of China (No. 62376006). We also thank Yang Liu and Zhuoqun Xu from Polar Mirage for their technical support on assets.
{
    \small
    \bibliographystyle{ieeenat_fullname}
    \bibliography{main}
}

\end{document}